\documentclass{article}

\usepackage{arxiv}

\usepackage[utf8]{inputenc} 
\usepackage[T1]{fontenc}    
\usepackage{hyperref}       
\usepackage{url}            
\usepackage{booktabs}       
\usepackage{amsfonts}       
\usepackage{nicefrac}       
\usepackage{microtype}      
\usepackage{lipsum}         
\usepackage{graphicx}
\usepackage{natbib}
\usepackage{doi}

\usepackage[american]{babel}

\usepackage{natbib} 
\usepackage{mathtools} 
\usepackage{booktabs} 
\usepackage{tikz} 

\usepackage{microtype}
\usepackage{graphicx}
\usepackage{subfigure}
\usepackage{subcaption}
\usepackage[ruled,noresetcount,vlined,linesnumbered]{algorithm2e}
\SetKwInput{KwInput}{Input}                
\SetKwInput{KwOutput}{Output}     
\usepackage{enumitem}

\usepackage{booktabs} 

\usepackage{hyperref}
\usepackage{bm} 

\usepackage{nicefrac }

\usepackage{amsmath}
\usepackage{amssymb}
\usepackage{mathtools}
\usepackage{amsthm}
\usepackage{cleveref}       

\theoremstyle{plain}

\theoremstyle{definition}

\theoremstyle{remark}

\DeclareMathOperator*{\argmax}{argmax}

\newcommand{\rev}[1]{{\color{purple}{#1}}
	\todo[caption={},color=purple!20!]
		{{\footnotesize to review}}
	}

 \newcommand{\cB}{\mathcal{B}} 
\newcommand{\cC}{\mathcal{C}} 
\newcommand{\cE}{\mathcal{E}} 
 
 \newcommand{\cL}{\mathcal{L}}
\newcommand{\cM}{\mathcal{M}} 
\newcommand{\cO}{\mathcal{O}} \newcommand{\cP}{\mathcal{P}}
 \newcommand{\cS}{\mathcal{S}}
 
\newcommand{\cZ}{\mathcal{Z}}

 \newcommand{\RR}{\mathbb{R}}

\DeclareMathOperator{\proj}{proj}

\newcommand{\maximize}[1]{\underset{{#1}}{\text{maximize}}}

\usepackage{esvect}

\usepackage[textsize=tiny]{todonotes}

\title{End-to-End Learning for Fair Multiobjective Optimization Under Uncertainty}

\author{ 
\href{https://orcid.org/0000-0002-6367-9626}{\includegraphics[scale=0.06]{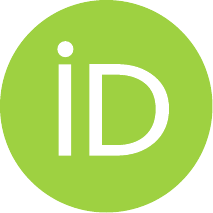}\hspace{1mm}My H.~Dinh} \\
	University of Virginia\\
	Charlottesville, VA, USA\\
	\texttt{fqw2tz@virginia.edu}\\
	\And
	\href{https://orcid.org/0000-0002-4499-8066}
    {\includegraphics[scale=0.06]{orcid.pdf}\hspace{1mm}
    James Kotary} \\
	University of Virginia\\
	Charlottesville, VA, USA\\
	\texttt{jk4pn@virginia.edu} \\
    \And
     \href{https://orcid.org/0000-0002-1381-6776}{\includegraphics[scale=0.06]{orcid.pdf}\hspace{1mm}Ferdinando Fioretto
     } \\
	University of Virginia\\
	Charlottesville, VA, USA\\
	\texttt{fioretto@virginia.edu} \\
}

\renewcommand{\shorttitle}{End-to-End Learning for Fair Multiobjective Optimization Under Uncertainty}

\begin{document}
\maketitle

\begin{abstract}
Many decision processes in artificial intelligence and operations research are modeled by parametric optimization problems whose defining parameters are unknown and must be inferred from observable data. The Predict-Then-Optimize (PtO) paradigm in machine learning aims to maximize downstream decision quality by training the parametric inference model end-to-end with the subsequent constrained optimization. This requires backpropagation through the optimization problem using approximation techniques specific to the problem's form, especially for nondifferentiable linear and mixed-integer programs. This paper extends the PtO methodology to optimization problems with nondifferentiable Ordered Weighted Averaging (OWA) objectives, known for their ability to ensure properties of fairness and robustness in decision models. Through a collection of training techniques and proposed application settings, it shows how optimization of OWA functions can be effectively integrated with parametric prediction for fair and robust optimization under uncertainty.
\end{abstract}

\keywords{Predict-then-Optimize \and Multi-Objectives Optimization \and Fairness}

\section{Introduction}
\label{sec:introduction}
The \emph{Predict-Then-Optimize} (PtO) framework models decision-making processes as optimization problems with unspecified parameters $\bm{c}$, which must be estimated by a machine learning (ML) model, given correlated features $\bm{z}$. An estimation of $\bm{c}$ completes the problem's specification, whose solution defines a mapping: 
\begin{equation}
    \label{eq:opt_generic}
    \bm{x}^\star(\bm{c}) = \argmax_{\bm{x} \in \cS} \;\; f(\bm{x}, \bm{c})
\end{equation}
The goal is to learn a model $\hat{\bm{c}} = \cM_{\theta}(\bm{z})$ 
from observable features $\bm{z}$, such that the objective value $f(\bm{x}^{\star}(\hat{\bm{c}}),\bm{c})$ under ground-truth parameters $\bm{c}$ is maximized on average. 

This setting is common to many real-world applications requiring decision-making under uncertainty, such as planning the fastest route through a city with unknown traffic delays, or determining optimal power generation schedules based on demand forcasts. A classic example is the Markowitz portfolio problem, wherein the optimization model \eqref{eq:opt_generic} may regard $f$ as the total return due to asset allocations $\bm{x}$ under predicted prices $\bm{c}$, while $\cS$ includes constraints on price covariance as a measure of risk \citep{markowitz1991foundations}. Modern approaches are based on \emph{end-to-end learning}, and train $\hat{\bm{c}} = \cM_{\theta}(\bm{z})$ to maximize $f(\bm{x}^{\star}(\hat{\bm{c}}),\bm{c})$ directly as loss function. This requires backpropagation through $\bm{x}^{\star}(\hat{\bm{c}})$, which is challenging when \eqref{eq:opt_generic} defines a nondifferentiable mapping, as it will be further elaborated in Section \ref{sec:preliminaries}.

Within this context, optimizing multiple objectives becomes an important extension, where the decision-making process needs to balance various competing objectives. 
Of particular interest is the case when such objectives must be \emph{fairly} optimized, as common in many engineering settings including energy systems \citep{TERLOUW2019356},  urban planning \citep{Salas2020Enhancing}, and multi-objective portfolio optimization \citep{Iancu2014Fairness,Chen2022Multiobjective}.  
In this setting, a prevalent approach is based on optimization of the scalar aggregation of all objectives by Ordered Weighted Averaging (OWA) \citep{Yager199380}. Such approach results in Pareto-optimal solutions that fairly balance the values of each individual objective. 
However, employing an OWA optimization within a Predict-Then-Optimize framework is challenged by its nondifferentiability, which prevents  backpropagation of its constrained optimization mapping $\bm{x}^{\star}(\bm{c})$ within machine learning models trained by gradient descent.

This paper aims to solve this challenge, 
and enable the combined learning and optimization of new applications such as fair learning-to-rank models based on OWA optimization of rankings, and Markowitz prediction models based on multiscenario portfolio optimization.  By leveraging modern techniques in OWA optimization and Predict-Then-Optimize learning, this paper shows how the optimization of OWA functions can be effectively backpropagated in machine learning models, enabling end-to-end trainable prediction and decision models for applications requiring fair and robust decision-making under uncertainty. 

\textbf{Contributions.} In particular, the paper makes the following contributions: \textbf{(1)} It proposes novel techniques for differentiating OWA optimization models with respect to their uncertain parameters, allowing their integration in end-to-end trainable ML models. \textbf{(2)} It is the first to show how loss functions based on OWA aggregation can be effectively employed for supervision of such end-to-end training. \textbf{(3)} Based on these contributions, it proposes several effective modeling strategies for combining parametric prediction with OWA optimization,  and evaluates them on novel application settings in which optimal decisions must be made fair or robust to multiple uncertain objective criteria.

\section{Preliminaries}
\label{sec:preliminaries}
Prior to discuss the paper contribution, this section provides an overview of the concepts of optimizing OWA functions and implementing end-to-end training methods for both prediction and optimization. 

\vspace{-6pt}
\subsection{OWA and its Optimization}
The $\textit{Ordered Weighted Average}$ (OWA) operator \citep{Yager199380} is a class of functions meant for aggregating multiple independent values, in settings requiring multicriteria evaluation and comparison \citep{10.5555/2464909}. Let $\mathbf{y} \in \mathbb{R}^m$ be a vector of $m$ distinct criteria, and $\tau: \mathbb{R}^m \rightarrow \mathbb{R}^m$ be the sorting map for which $\tau(\mathbf{y}) \in \mathbb{R}^m$ holds the elements of $\mathbf{y}$ in increasing order. Then for any $\mathbf{w}$ satisfying $\{ \mathbf{w} \in \mathbb{R}^m: \sum_i w_i = 1, \mathbf{w} \geq 0 \}$, the OWA aggregation with weights $\mathbf{w}$ is defined as a linear functional on  $\tau(\mathbf{y})$:
\begin{equation}
\label{eq:OWA_definition}
\textsc{OWA}_{\mathbf{w}}(\mathbf{y}) = \mathbf{w}^T \tau(\mathbf{y}),
\end{equation}
which is concave and piecewise-linear in $\mathbf{y}$ \citep{ogryczak2003solving}. 

\textbf{Fair OWA.}
Of particular interest are the \emph{Fair OWA}, whose weights have decreasing order: $w_1 > w_2 \ldots > w_n$. 

The following three properties possessed by Fair OWA functions are key to their use in fairly optimizing multiple objectives: 
{\bf (1)} Let $\cP_{m}$ be the set of all permutations of $[1,\ldots,m]$. \emph{Impartiality} means that Fair OWA treats all criteria equally, in the sense that $\textsc{OWA}_{\mathbf{w}}(\mathbf{y}) = \textsc{OWA}_{\mathbf{w}}(\mathbf{y}_{\sigma})$ for any $\sigma \in \cP_{m}$. 
{\bf (2)} \emph{Equitability} is the property that marginal transfers from a criterion with higher value to one with lower value results in an increased OWA aggregated value. This condition holds that $\textsc{OWA}_{\mathbf{w}}(\mathbf{y}_\epsilon) >  \textsc{OWA}_{\mathbf{w}}(\mathbf{y})$, where $\mathbf{y}_{\epsilon} = \mathbf{y}$ except at positions $i$ and $j$ where $(\mathbf{y}_{\epsilon})_i = \mathbf{y}_i - \epsilon$ and $(\mathbf{y}_{\epsilon})_j = \mathbf{y}_j + \epsilon$, assuming  $y_i > y_j +\epsilon$.  
{\bf (3)} \emph{Monotonicity} means that $\textsc{OWA}_{\mathbf{w}}(\mathbf{y})$ is an increasing function of each element of $\mathbf{y}$. The monotonicity property implies that solutions which optimize \eqref{eq:OWA_definition} are Pareto Efficient solutions of the underlying multiobjective problem, thus that no single criteria can be raised without reducing another \cite{ogryczak2003solving}. 
Intuitively, OWA objectives lead to fair optimal solutions by always assigning the highest weights of $\bm{w}$ to the objective criteria in order of lowest current value.

\vspace{-6pt}
\subsection{Predict-Then-Optimize Learning}
\label{subsec:predict-then-optimize}

The problem setting of this paper can be viewed within the framework of Predict-Then-Optimize. Generically, a parametric optimization problem \eqref{eq:opt_generic} models an optimal decision $\bm{x}^{\star}(\bm{c})$ with respect to unknown parameters $\bm{c}$ within a distribution $\bm{c} \sim \cC$. Although the true value of $\bm{c}$ is unknown, correlated \emph{feature} values $\bm{z} \sim \cZ$ can be observed. The goal is to learn a predictive model $\cM_{\theta}: \cZ \to \cC$ from features $\bm{z}$ to estimate problem parameters $\hat{\bm{c}} = \cM_{\theta}(\bm{z})$, such that the resulting solution's  empirical objective value under ground-truth parameters, is maximized. That is, 
\begin{equation}
    \label{eq:OWA_PtO_ERM}
         \argmax_{\theta} \;\;\; \mathbb{E}_{(\bm{z},\bm{c}) \sim \Omega} \; f \left( \bm{x}^{\star}(\cM_{\theta}(\bm{z})), \bm{c} \right),
\end{equation}
where $\Omega$ represents the joint distribution between $\cZ$ and $\cC$. 

The above training goal is often best realized by maximizing empirical \emph{Decision Quality} as a loss function \cite{mandi2023decision}, defined
\begin{equation}
    \label{eq:DQ_loss}
    \cL_{DQ}(\hat{\bm{c}},\bm{c}) = f \left( \bm{x}^{\star}(\hat{\bm{c}}), \bm{c} \right).
\end{equation}
Gradient descent training of \eqref{eq:OWA_PtO_ERM} with $\cL_{DQ}$ requires a model of gradient $\frac{\partial \cL_{DQ}}{\partial \hat{\bm{c}}}$, either directly or through chain-rule composition $\frac{\partial \cL_{DQ}}{\partial \hat{\bm{c}}} = \frac{\partial  \bm{x}^{\star}(\hat{\bm{c}})}{\partial \hat{\bm{c}}} \cdot \frac{\partial \cL_{DQ}}{\partial \bm{x}^{\star}}$. Here, left-multiplication by the Jacobian is equivalent to backpropagation through the optimization mapping $\bm{x}^{\star}$. When $\bm{x}^{\star}$ is not differentiable, as in the case of OWA optimizations, smooth approximations are required, such as those developed in the next section. 

\begin{figure*}[ht]
\center
\includegraphics[width=1.0\columnwidth]{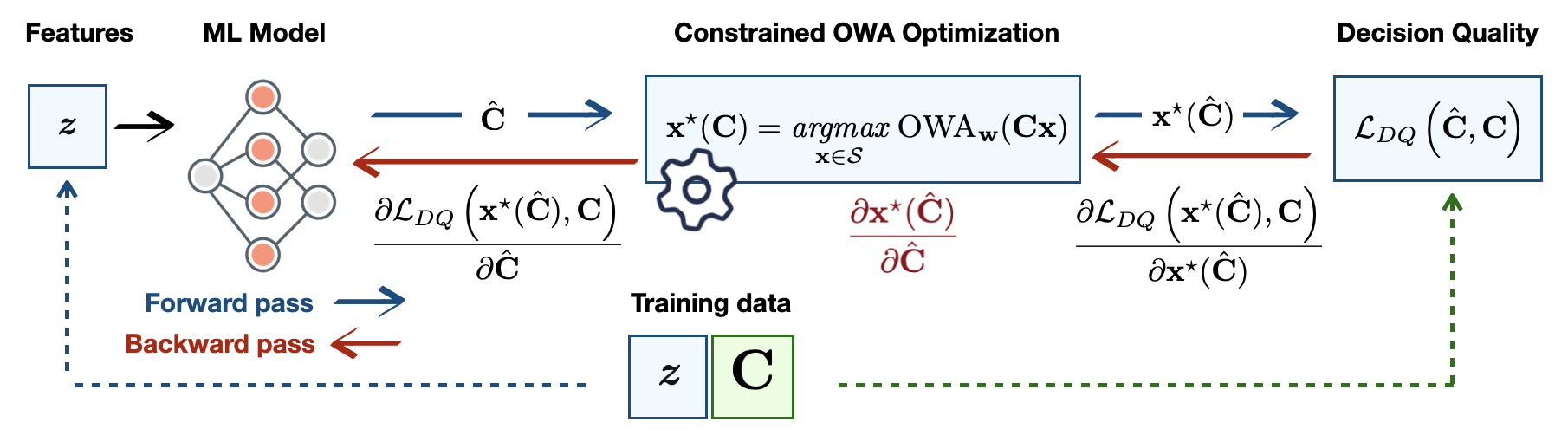}

\caption{Predict-Then-Optimize for OWA Optimization.}

\label{fig:scheme}

\end{figure*}


\section{End-to-End Learning with Fair OWA Optimization}
\label{sec:learning_with_OWA}

This paper's proposed methodology and setting are focused on when the objective function $f$ is composed of an ordered weighted average of $m$ linear objective functions,  each parametrized by one row of a matrix $\bm{C} \in \mathbb{R}^{m \times n}$, so that $f(\bm{x},\bm{C}) = \textsc{OWA}_{\bm{w}}(\bm{C}\bm{x})$ and
\begin{equation}
\label{eq:param_OWA_general}
    \bm{x}^\star(\bm{C}) = \argmax_{\bm{x} \in \cS} \;\;  \textsc{OWA}_{\bm{w}}(\bm{C}\bm{x}).
\end{equation}
Note that the methodology of this paper naturally extends  to cases where the OWA objective above is combined with additional smooth objective terms. For simplicity, the exposition is developed primarily in terms of the pure OWA objective as shown in equation \eqref{eq:param_OWA_general},  wherever applicable.

The goal is to learn a prediction model $\hat{\bm{C}} = \cM_{\theta}(\bm{z})$ that maximizes decision quality through gradient descent on problem \eqref{eq:OWA_PtO_ERM}, which requires its gradients w.r.t. $\hat{\bm{C}}$:
\begin{equation}
\label{eq:OWA_grad_comp}
    \frac{\partial \cL_{DQ}(\hat{\bm{C}},\bm{C})}{\partial \hat{\bm{C}}}  = \underbrace{\frac{\partial \bm{x}^{\star}}{\partial \hat{\bm{C}}}}_{\bm{J}} \; \cdot \underbrace{\frac{\partial \textsc{OWA}_{\bm{w}}(\bm{C} \bm{x}^{\star})}{\partial \bm{x}^{\star}}}_{\bm{g}},  
\end{equation}
where $\bm{x}^{\star}$ is evaluated at $\hat{\bm{C}}$. The primary strategy for modeling this overall gradient involves initially determining the OWA function gradient $\bm{g}$, followed by computing the product $\bm{J}\bm{g}$ by backpropagation of $\bm{g}$ through $\bm{x}^{\star}$.  

While nondifferentiable, the class of OWA functions is \emph{subdifferentiable}, with subgradients  as follows:
\begin{equation}
    \label{eq:owa_subgrad}
    \frac{\partial}{\partial \bm{y}} \textsc{OWA}_{\bm{w}}(\bm{y}) = \bm{w}_{(\sigma^{-1})} 
\end{equation}
where $\sigma$ are the sorting indices on  $\bm{y}$ \citep{do2022optimizing}. Based on this formula, computing an overall subgradient $\bm{g} = \nicefrac{\partial}{\partial \bm{x}} \, \textsc{OWA}_{\bm{w}}(\bm{C}\bm{x})$ is a routine application of the chain rule (via automatic differentiation). While the subgradients \eqref{eq:owa_subgrad} have been used in OWA optimization, this is the first work which demonstrates their use in training ML models. 
A schematic illustration highlighting the forward and backward steps required for this process is provided in Figure~\ref{fig:scheme}.


As outlined next, the main technical contribution of the paper is to propose differentiable models of OWA optimization \eqref{eq:param_OWA_general}, through which backpropagation of $\bm{g}$ can effectively approximate the decision quality gradient $\bm{J} \bm{g}$ for end-to-end training of \eqref{eq:OWA_PtO_ERM}. The following sections propose alternative models of differentiable OWA optimization, each taylored to address problem-specific technical challenges.

First, Section \ref{sec:diff_owa_opt} shows how the OWA optimization \eqref{eq:param_OWA_general} with continuous variables can be effectively smoothed, to yield differentiable approximations that can be backpropagated in end-to-end training \eqref{eq:OWA_PtO_ERM}. Then, Section  \ref{subsec:nonparametric_OWA_LP} focuses on a special form of  optimization mapping with nonparametric OWA term but an additional parametric objective term, and shows how its backpropagation can implemented using only a blackbox solver of the underlying problem, without smoothing. Finally Section \ref{subsec:surrogate_projection} outlines a method of surrogate solvers, focusing on cases where OWA-aggregation of objectives renders its optimization too difficult to solve. 

\section{Differentiable Approximate OWA Optimization}
\label{sec:diff_owa_opt}

This section develops two alternative differentiable approximations of the OWA optimization mapping \eqref{eq:param_OWA_general}. 
Prior works \citep{wilder2019melding,amos2019limited} show that when an optimization mapping \eqref{eq:opt_generic} is discontinuous, as is the case when $f$ and $\cS$ define a linear program (LP), differentiable approximations to \eqref{eq:opt_generic} can be formed by regularization of its objective by smooth functions. Section \ref{subsec:OWA_LP_quadreg} will show how linear programming  models of OWA optimization can be combined with smoothing techniques for LP, which perform well as differentiable approximations of \eqref{eq:param_OWA_general}. 

However, this model is shown to become computationally intractable for more than a few criteria $m$. An efficient alternative is proposed in Section \ref{subsec:moreau_solver}, in which the mapping \eqref{eq:param_OWA_general} is made differentiable by replacing the OWA objective with its smooth Moreau envelope approximation. To the best of the authors knowledge, this is the first time that objective smoothing via the Moreau envelope is used (and shown be an effective technique) for approximating nondifferentiable optimization programs in end-to-end learning. As approximations of the true mapping \eqref{eq:param_OWA_general}, both smoothed models are used employed in training and replaced by \eqref{eq:param_OWA_general} at test time, similarly to a softmax layer in classification.

\vspace{-6pt}
\subsection{OWA LP with Quadratic Smoothing}
\label{subsec:OWA_LP_quadreg}
The mainstay approach to solving problem \eqref{eq:param_OWA_general}, when $\bm{x} \in \cS$ is linear, is to transform the problem into a linear program without OWA functions, and solve it with a simplex method \citep{ogryczak2003solving}.
Our first approach to differentiable OWA optimization combines this transformation with the smoothing technique of \cite{wilder2019melding}, which forms differentiable approximations to linear programs
\begin{equation}
    \label{model:LP_generic}
    \bm{x}^{\star}(\bm{c}) = {\argmax}_{ \bm{A}\bm{x} \leq \bm{b}} \;\; \bm{c}^T \bm{x}  
\end{equation}
by adding a scaled euclidean norm term $\epsilon \| \bm{x} \|^2$ to the objective function, resulting in a continuous mapping $ \bm{x}^{\star}(\bm{c}) = {\argmax}_{ \bm{A}\bm{x} \leq \bm{b}} \;\; \bm{c}^T \bm{x} + \epsilon \| \bm{x} \|^2  $, a quadratic program (QP) which can be differentiated implicitly via its KKT conditions as in  \citep{amos2017optnet}.

We adapt a version of this technique to OWA optimization \eqref{eq:param_OWA_general} by first forming an equivalent LP problem. It is observed in \citep{ogryczak2003solving} that $\textsc{OWA}_{\bm{w}}$ can be expressed as the minimum weighted average resulting among all permutations of the OWA weights $\bm{w}$:
\begin{equation}
    \label{model:OWA_LP_form}
         \textit{OWA}_{\mathbf{w}}(\mathbf{r}) = {\max}_{z} \;\; z \;\;\; \textit{s.t.} \;\;\;\; z \leq \mathbf{w}_{\sigma} \cdot \mathbf{r}, \;\;\; \forall \sigma \in \cP, 
\end{equation}
which allows the OWA optimization \eqref{eq:param_OWA_general} to be expressed as
\begin{subequations}
    \label{model:OWA_LP_opt}
    \begin{align}
    \label{model:OWA_LP_opt_obj}
    \bm{x}^{\star}(\bm{C}) = {\argmax}_{ \bm{x} \in \cS, \bm{y},z} &\;\; 
     z  \\
     \label{model:OWA_LP_opt_obj_constr1}
    \mbox{s.t.:} &\;\; \bm{y} = \bm{C}\bm{x} \\
    \label{model:OWA_LP_opt_obj_constr2}
    &\;\; z \leq \bm{w}_{\tau} \cdot \bm{y} \;\;\; \forall \tau \in \mathcal{P}_m.
    \end{align}
\end{subequations}
When the constraints $\bm{x} \in \cS$ are linear, problem \eqref{model:OWA_LP_opt} is a LP. However, its constraints \eqref{model:OWA_LP_opt_obj_constr2} grow factorially as $m!$, where $m$ is the number of individual objective criteria aggregated by OWA. Smoothing by the scaled norm of joint variables $\bm{x},\bm{y},z$ leads to a differentiable QP approximation, viable when $m$ is small, which can be solved and differentiated using \citep{amos2017optnet} or the generic differentiable optimization solver \citep{agrawal2019differentiable}:
\begin{subequations}
    \label{model:OWA_LP_smooth}
    \begin{align}
    \label{model:OWA_LP_smooth_obj}
    \bm{x}^{\star}(\bm{C}) = \argmax_{\substack{\bm{x} \in \cS, \bm{y}, z}} &\;\;
     z + \epsilon \left( \|\bm{x}\|_2^2  + \|\bm{y}\|_2^2 + z^2 \right)  \\
    \mbox{subject to:} & \;\;\; \eqref{model:OWA_LP_opt_obj_constr1},\eqref{model:OWA_LP_opt_obj_constr2}.
    \end{align}
\end{subequations}
While problem \eqref{model:OWA_LP_opt} does not fit the exact form \eqref{model:LP_generic} due to its parameterized constraints \eqref{model:OWA_LP_opt_obj_constr1}, the need for quadratic smoothing \eqref{model:OWA_LP_smooth_obj} is illustrated experimentally in Section \ref{sec:portfolio}.
The main \emph{disadvantage} of this method is poor scalability in the number of criteria $m$, due to  constraints \eqref{model:OWA_LP_opt_obj_constr2}.

\vspace{-6pt}
\subsection{Moreau Envelope Smoothing}
\label{subsec:moreau_solver}

In light of the efficiency challenges faced by \eqref{model:OWA_LP_smooth}, we propose an alternative smoothing technique to form more scalable differentiable approximations of the optimization mapping \eqref{eq:param_OWA_general}.  Rather than adding a quadratic term as in \eqref{model:OWA_LP_smooth}, we replace the piecewise linear function $\textsc{OWA}_{\bm{w}}$ in \eqref{eq:param_OWA_general} by its Moreau envelope, defined for generic $f$ as:
\begin{equation}
    \label{eq:moreau_env}
    f^{\beta}(\mathbf{x}) = \min_{\mathbf{v}} \;\;  f(\mathbf{v}) + \frac{1}{2 \beta} \|   \mathbf{v} - \mathbf{x}  \|^2.
\end{equation}
Compared to its underlying function $f$, the Moreau envelope is $\frac{1}{\beta}$ smooth while sharing the same optima \citep{beck2017first}. The Moreau envelope-smoothed OWA optimization is
\begin{equation}
    \label{model:OWA_moreau_solver}
    \bm{x}^{\star}(\bm{C}) = {\argmax}_{\bm{x} \in \cS} \;\; 
     \textsc{OWA}_{\bm{w}}^{\beta}(\bm{C}\bm{x}). 
\end{equation}
With its smooth objective function, problem \eqref{model:OWA_moreau_solver} can be solved by gradient-based optimization methods, such as projected gradient descent, or more likely a Frank-Wolfe method if $\bm{x} \in \cS$ is linear (see Section \ref{sec:portfolio}). Additionally, it can be effectively backpropagated in end-to-end learning.

Backpropagation of \eqref{model:OWA_moreau_solver} is nontrivial since its objective function lacks a closed form. To proceed, we first note from \citep{do2022optimizing} that the gradient of the Moreau envelope $\textsc{OWA}_{\bm{w}}^{\beta}$  is equal to a Euclidean projection:
\begin{equation}
    \label{eq:OWA_moreau_grad}
    \frac{\partial}{\partial \bm{x}} \textsc{OWA}_{\bm{w}}^{\beta}(\mathbf{x}) = \proj_{\cC(\tilde{\bm{w}})} \left(\frac{\bm{x}}{\beta}\right),
\end{equation}
where $\tilde{\bm{w}} = -(w_m,\ldots,w_1)$  and the permutahedron $\cC(\tilde{\bm{w}})$ is the convex hull of all permutations of $\tilde{\bm{w}}$.
It's further shown in \citep{blondel2020fast} how such a projection can be computed and differentiated in $\cO(m \log m)$ time using isotonic regression. To leverage the differentiable gradient function \eqref{eq:OWA_moreau_grad} for backpropagation of the smoothed optimization \eqref{model:OWA_moreau_solver},  we model its Jacobian by differentiating the fixed-point conditions of a gradient-based solver. 

Letting $\mathcal{U}(\bm{x},\bm{C}) =  \textit{proj}_{\cS}( \bm{x} - \alpha \cdot \frac{\partial}{\partial \bm{x}} \textsc{OWA}_{\bm{w}}^{\beta}(\bm{x}, \bm{C} ))$, a projected gradient descent step on \eqref{model:OWA_moreau_solver} is $\bm{x}^{k+1} = \mathcal{U}(\bm{x}^k,\bm{C})$. Differentiating the fixed-point conditions of convergence where $\bm{x}^{k} = \bm{x}^{k+1} = \bm{x}^{\star}$, and rearranging terms yields a linear system for $\frac{\partial \bm{x}^{\star}}{\partial \bm{C}}$:

\begin{equation}
    \label{eq:PGD_grad}
       \left( I - \underbrace{ \frac{\partial \mathcal{U}(\bm{x}^{\star},\bm{C})}{\partial \bm{x}^{\star}} }_{\bm{\Phi}} \right) \frac{\partial \bm{x}^{\star}}{\partial \bm{C}} =  \underbrace{ \frac{\partial \mathcal{U}(\bm{x}^{\star},\bm{C})}{\partial \bm{C}} }_{\bm{\Psi}}
\end{equation}
The partial Jacobian matrices $\bm{\Phi}$ and  $\bm{\Psi}$ above can be found given a differentiable implementation of $\mathcal{U}$. This is achieved by computing the inner gradient $\frac{\partial}{\partial \bm{x}} \textsc{OWA}_{\bm{w}}^{\beta}(\bm{x}, \bm{C} )$ via the differentiable permutahedral projection \eqref{eq:OWA_moreau_grad}, and solving the outer projection mapping $\proj_{\cS}$ using a generic differentiable solver such as \texttt{cvxpy} \citep{agrawal2019differentiable}. As such, applying $\mathcal{U}$ at a precomputed solution $\bm{x}^{\star}(\bm{C})$ allows $\bm{\Phi}$ and  $\bm{\Psi}$ to be extracted in PyTorch, in order to solve \eqref{eq:PGD_grad}; this process is efficiently implemented via the \texttt{fold-opt} library \citep{kotary2023folded}.

\section{Blackbox Methods for Nonparametric OWA Objective}
\label{subsec:nonparametric_OWA_LP}
This section proposes a special class of techniques for cases where the OWA term of an objective function is specified with \emph{known} coefficients $\bm{B} \in \mathbb{R}^{m \times n}$, and uncertainty lies instead in an additional parametrized \emph{linear} objective term: 
\begin{equation}
    \label{model:OWA_opt_nonparametric}
    \bm{x}^{\star}(\bm{c}) = {\argmax}_{\bm{x} \in \cS} \;\; 
     \bm{c}^T \bm{x} + \lambda \textsc{OWA}_{\bm{w}}(\bm{B}\bm{x}).
\end{equation}
This form is taken by the optimization mapping within the fair learning to rank model proposed in Section \ref{sec:fair_ltr}. Appealing again to the reformulation \eqref{model:OWA_LP_opt} leads \eqref{model:OWA_opt_nonparametric} to become
\begin{subequations}
    \label{model:OWA_LP_opt_plus_linear}
    \begin{align}
    \left(\bm{x}^{\star},\bm{y}^{\star},z^{\star}\right)(\bm{c}) = 
    &{\argmax}_{ \bm{x} \in \cS, \bm{y},z} \; 
     \bm{c}^T \bm{x}  +  \lambda z  \\
    \mbox{s.t.:} &\;\; \bm{y} = \bm{B}\bm{x} \\
    \label{model:OWA_LP_opt_plus_linear_obj_constr2}
    &\;\; z \leq \bm{w}_{\tau} \cdot \bm{y} \;\;\; \forall \tau \in \mathcal{P},
    \end{align}
\end{subequations}
which as discussed in Subsection \ref{subsec:OWA_LP_quadreg} grows intractable with increasing $m$ since the constraints \eqref{model:OWA_LP_opt_plus_linear_obj_constr2} number $(m!)$. 

On the other hand, it fits the particular form $\bm{v}^{\star}(\bm{\gamma}) = \argmax_{\bm{v} \in \cC} \; \bm{\gamma}^T \bm{v}$, treated in several works \citep{elmachtoub2020smart,berthet2020learning,vlastelica2020differentiation}, wherein an uncertain \emph{linear} objective is paired with \emph{nonparametric} constraints. These works propose differentiable solvers based on \emph{blackbox} solvers of the underlying optimization problem, without smoothing. This is generally accomplished by modeling the gradient as a combination of solutions induced by perturbed input parameters. As shown next, this allows a gradient formula based on \eqref{model:OWA_LP_opt_plus_linear} to be computed without actually solving \eqref{model:OWA_LP_opt_plus_linear}, given a blackbox solver for the underlying problem \eqref{model:OWA_opt_nonparametric}.

We illustrate the idea using the "Smart Predict-Then-Optimize" scheme \citep{elmachtoub2020smart}, which trains to maximimize $\cL_{DQ}$ by equivalently minimizing the suboptimality (called \emph{regret}) via a convex subdifferentiable upper bounding function named $\cL_{SPO+}$. By construction it admits a formula for subgradients directly with respect to $\hat{\bm{c}}$:
\begin{equation}
\label{model:SPO_grad}
 \nicefrac{\partial}{\partial \hat{\bm{\gamma}}}\; L_{\texttt{SPO}+}(\hat{\bm{\gamma}},\bm{\gamma}) =\bm{v}^{\star}(2\hat{\bm{\gamma}} - \bm{\gamma})  - \bm{v}^{\star}(\bm{\gamma})  .
\end{equation}
Given any efficient method which provides optimal solutions $\bm{x}^{\star}(\bm{C})$ to the OWA optimization \eqref{model:OWA_opt_nonparametric}, the auxiliary variables of problem \eqref{model:OWA_LP_opt_plus_linear} can be recovered as $\bm{y}^{\star} = \bm{C}\bm{x}^{\star}$ and $z^{\star} = \textsc{OWA}_{\bm{w}}(\bm{y}^{\star})$. Defining the variables $\bm{v}^{\star} = (\bm{x}^{\star},\bm{y}^{\star},z^{\star})$ and noting that $\bm{\gamma} = (\bm{c},\bm{0},\lambda)$ in problem \eqref{model:OWA_LP_opt_plus_linear}, its SPO+ loss subgradient can be now computed directly using formula \eqref{model:SPO_grad}. In this way, the problem form \eqref{model:OWA_LP_opt_plus_linear} is leveraged to derive a \emph{backpropagation model}, while \emph{avoiding its direct solution} as a linear program. Section \ref{sec:portfolio},  will show how this can be applied in combination with an efficient Frank-Wolfe solution of \eqref{model:OWA_opt_nonparametric} to design a scalable fair learning to rank model.

\section{Differentiable Surrogate Optimization Mappings}
\label{subsec:surrogate_projection}

OWA optimization problems \eqref{eq:param_OWA_general} can be difficult to solve in general, even with modern methods, without exploiting special problem-specific structure. In such cases, an alternative to differentiable approximations of \eqref{eq:param_OWA_general}, as proposed in Section \ref{sec:diff_owa_opt}, would be to produce feasible candidate $\bm{x}^{\star} \in \cS$ from a simpler differentiable model without OWA objectives. 

For example, a \emph{linear} surrogate model proves useful when \eqref{eq:param_OWA_general} represents fair OWA optimization of multiple objectives in a linear program (such as shortest path or bipartite matching) which depends on total unimodularity to maintain integral solutions $\bm{x}^{\star} \in \{ 0,1 \}^n$:
\begin{equation}
    \label{model:surrogate_LP}
    \bm{x}^{\star}(\bm{c}) = {\argmax}_{ \bm{x} \in \cS} \;\;  \bm{c}^T  \bm{x}
\end{equation}
As illustrated in Section \ref{sec:experiments_warcraft} on a parametric shortest path problem, this surrogate approach is essential to avoid arising an intractable OWA mixed-integer program, since integrality of solutions is guaranteed only under linear objectives. 

The main \emph{disadvantage} inherent to the proposed surrogate models is that they do not directly approximate the true OWA problem \eqref{eq:param_OWA_general}. Thus, the learned model $ \cM_{\theta}(\bm{z}) = \hat{\bm{c}}  \in \mathbb{R}^n$ does not fit the form prescribed in (\eqref{eq:OWA_PtO_ERM},\eqref{eq:param_OWA_general})  as written, and it cannot supply parametric estimates $\hat{\bm{C}} \in \mathbb{R}^{m \times n}$ to an  external solver of problem \eqref{eq:param_OWA_general}. Despite this, using $\textsc{OWA}_{\bm{w}}(\bm{C}\bm{x}(\hat{\bm{c}}))$ as a loss function trains the surrogate model to learn solutions to \eqref{eq:param_OWA_general} with high decision quality.

\begin{figure*}[!t]
\centering
\includegraphics[width=1\columnwidth]{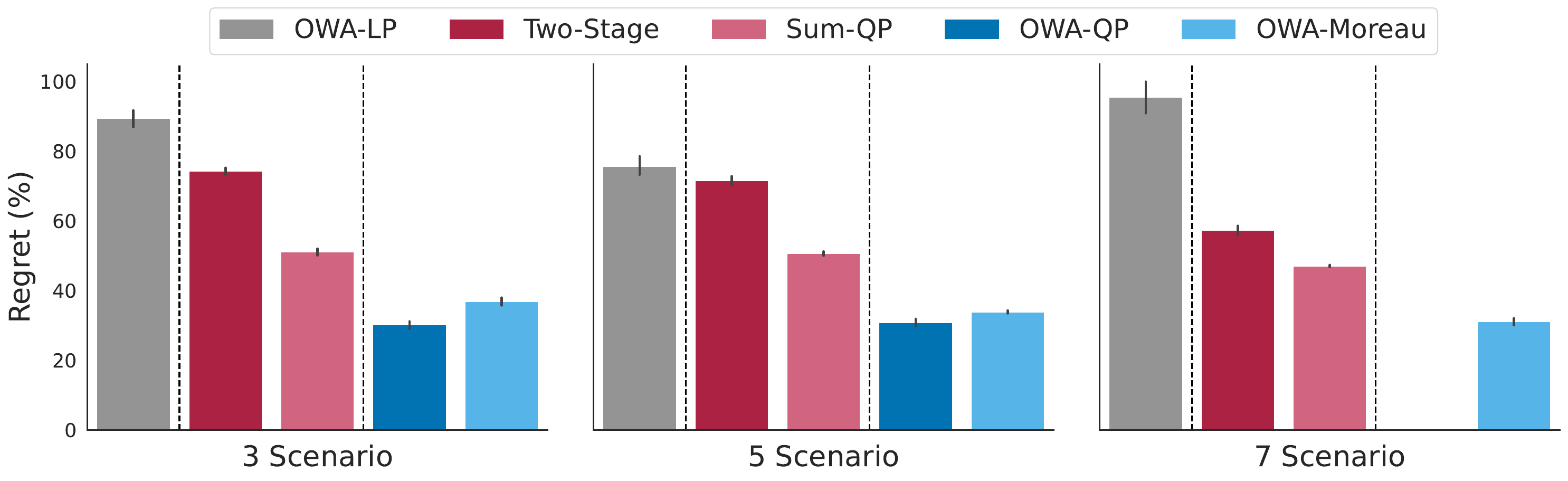}
\caption{Percentage OWA regret (lower is better) on test set,  on robust portfolio problem over $3$,$5$,$7$ scenarios. }
\label{fig:barcharts_portfolio}
\end{figure*}

\section{Experiments}
\label{sec:experiments}
This section uses the differentiable elements introduced in Sections \ref{sec:diff_owa_opt}-\ref{subsec:surrogate_projection} to compose end-to-end trainable prediction and OWA optimization models. Three experimental settings are proposed for their evaluation, across two main application settings. The first application setting is \emph{Fair Multiobjective Prediction and Optimization}, which extends the Predict-Then-Optimize setting of Section \ref{subsec:predict-then-optimize} to cases where multiple uncertain objective functions must be jointly learned and optimized fairly via their OWA aggregation, as per problem \eqref{eq:param_OWA_general}. Within this setting, \emph{Robust Markowitz Portfolio Optimization} focuses on comparatively evaluating the differentiable approximations proposed in Section \ref{sec:diff_owa_opt} against a host of baseline methods. Then, \emph{Multi-Species Warcraft Shortest Path} serves as a case study in which a differentiable surrogate model can be enable learning with OWA optimization of integer variables. A second, distinct application setting proposes a \emph{Fair Learning-to-Rank} model, whose OWA-aggregated objectives are known with certainty, corresponding to problem \eqref{model:OWA_opt_nonparametric} as detailed in Section \ref{subsec:nonparametric_OWA_LP}.

\vspace{-6pt}
\subsection{Fair Multiobjective Predict-and-Optimize}
This setting uses a prediction model $\hat{\bm{C}} = \cM_{\theta}(\bm{z})$ to jointly estimate from features $\bm{z}$ the coefficients $\bm{C} \in \mathbb{R}^{m \times n}$ of $m$ linear objectives, taken together as $\bm{C}\bm{x} \in \mathbb{R}^m$. Its training goal is to maximize empirical decision quality with respect to their Fair OWA aggregation $f(\bm{x},\bm{C}) = \textsc{OWA}_{\bm{w}}(\bm{C}\bm{x})$:
\begin{equation}
    \label{eq:DQ_loss_OWA}
    \cL_{DQ}(\hat{\bm{C}},\bm{C}) = \textsc{OWA}_{\bm{w}}\left(\bm{C}\bm{x}^{\star}(\hat{\bm{C}})\right).
\end{equation}
Any descending OWA weights $\bm{w}$ can be used to specify \eqref{eq:DQ_loss_OWA}; we choose the squared \emph{Gini indices} $w_j = \left( \frac{n-1+j}{n} \right)^2$.

\paragraph{Evaluation.} In this section, each model is evaluated on the basis of its ability to train a model $\hat{\bm{C}} = \cM_{\theta}(z)$ to attain high decision quality \eqref{eq:DQ_loss_OWA} in terms of the OWA-aggregated objective. Results are reported in terms of the equivalent \emph{regret} metric of suboptimality, whose minimimum value $0$ corresponds to maximum decision quality:
\begin{equation}
    \label{eq:regret}
    \textit{regret}(\hat{\bm{C}}, \bm{C}) = \textsc{OWA}^{\star}_{\bm{w}}\left(\bm{C}\right) - \textsc{OWA}_{\bm{w}}\left(\bm{C}\bm{x}^{\star}(\hat{\bm{C}})\right)
\end{equation}
where $\textsc{OWA}^{\star}_{\bm{w}}\left(\bm{C}\right)$ is the true optimal value of problem \eqref{eq:param_OWA_general}. This experiment is designed to evaluate the proposed differentiable approximations \eqref{model:OWA_LP_smooth} and \eqref{model:OWA_moreau_solver} of Section \ref{sec:diff_owa_opt}; for reference they are named \textit{OWA-QP} and \textit{OWA-Moreau}.

\paragraph{Baseline Models.} In addition to the newly proposed models, the evaluations presented in this section include two main baseline methods: \textbf{(1)} The \emph{two-stage} method is the standard baseline for comparison against proposed methods for Predict-Then-Optimize training \eqref{eq:OWA_PtO_ERM} \citep{mandi2023decision}. It trains the prediction model $\hat{\bm{C}} = \cM_{\theta}(\bm{z})$ by MSE regression, minimizing $\cL_{TS}(\hat{\bm{C}}, \bm{C}) = \| \hat{\bm{C}} - \bm{C} \|^2$ without considering the downstream optimization model, which is employed only at test time. In addition, \textbf{(2)} the \emph{unweighted sum} (\textit{UWS}) of the objective criteria results in an LP mapping $\bm{x}^{\star}(\bm{C}) = \argmax_{\bm{x} \in \cS} \bm{1}^T (\bm{C} \bm{x})$ which can be employed in end-to-end training by using quadratic smoothing \citep{wilder2019melding} in \ref{sec:portfolio} and blackbox differentiation \citep{vlastelica2020differentiation} in \ref{sec:experiments_warcraft}; this baseline leverages end-to-end learning but without incorporating the OWA objective.

\vspace{-6pt}
\subsubsection{Differentiable OWA Optimization: Robust Markowitz Portfolio Problem}
\label{sec:portfolio}
The classic Markowitz portfolio problem is concerned with constructing an optimal investment portfolio, given future returns $\bm{c} \in \mathbb{R}^n$ on $n$ assets, which are unknown and predicted from exogenous data. A common formulation maximizes future returns subject to a risk limit, modeled as a quadratic covariance constraint. Define the set of valid fractional allocations  $\Delta_n = \{\bm{x} \in \mathbb{R}^n: \bm{1}^T \bm{x} = 1, \bm{x}\geq 0 \}$, then :
\begin{equation}
    \label{eq:opt_portfolio_old}
        \bm{x}^\star(\bm{c}) = \argmax_{\bm{x} \in \Delta_n} \;\; \bm{c}^T \bm{x}  \;\;\;  
        \textsl{s.t.:} \;\;\; \bm{x}^T \bm{\Sigma} \bm{x} \leq \delta.
\end{equation}
where $\bm{\Sigma} \in \mathbb{R}^{n \times n}$ are the price covariances over $n$ assets. The  optimal portfolio allocation \eqref{eq:opt_portfolio_old} as a function of future returns $\bm{c} \in \mathbb{R}^n$ is differentiable using known methods \citep{agrawal2019differentiable}, and is commonly used in evaluation of Predict-Then-Optimize methods \citep{mandi2023decision}.

An alternative approach to risk-aware portfolio optimization views risk in terms of robustness over alternative scenarios. In \citep{cajas2021owa}, $m$ future price scenarios are modeled by a matrix $\bm{C} \in \mathbb{R}^{m \times n}$ whose $i^{\textit{th}}$ row holds per-asset prices in the $i^{\textit{th}}$ scenario. Thus an optimal allocation is modeled as
\begin{equation}
\label{eq:param_OWA_portfolio}
    \bm{x}^\star(\bm{C}) = \argmax_{\bm{x} \in \Delta_n} \;\;  \textsc{OWA}_{\bm{w}}(\bm{C}\bm{x}).
\end{equation}
This experiment integrates robust portfolio optimization \eqref{eq:param_OWA_portfolio} end-to-end with per-scenario price prediction $\hat{\bm{C}} = \cM_{\theta}(\bm{z})$. 

\paragraph{Settings.}
Historical prices of $n=50$ assets are obtained from the Nasdaq online database \citep{NASDAQ}  years 2015-2019, and $N = 5000$ baseline asset price samples $\bm{c}_i$ are generated by adding Gaussian random noise to randomly drawn price vectors. Price scenarios are simulated as a matrix of multiplicative factors uniformly drawn as $  \mathcal{U}(0.5,1.5)^{m \times n}$, whose rows are multiplied elementwise with $\bm{c}_i$ to obtain $\bm{C}_i \in \mathbb{R}^{m \times n}$. While future asset prices can be predicted on the basis of various exogenous data including past prices or sentiment analysis, this experiment generates feature vectors $\bm{z}_i$ using a randomly generated nonlinear feature mapping as described in Appendix \ref{app:portfolio}. The experiment is replicated in three settings which assume $m=3$, $5$, and $7$ scenarios.

The predictive model $\cM_{\theta}$ is a feedforward neural network. At test time, $\cM_{\theta}$ is evaluated over a test set for the distribution  $(\bm{z},\bm{C}) \in \Omega$, by passing its predictions to a projected subgradient solver of \eqref{eq:param_OWA_portfolio}. Complete details in Appendix \ref{app:portfolio}.

\paragraph{Results.} Figure \ref{fig:barcharts_portfolio} shows percent regret in the OWA objective attained on average over the test set (lower is better). The end-to-end trained unweighted sum baseline outperforms the two-stage, while notice how both \textit{OWA-QP} and \textit{OWA-Moreau} reach substantially higher decision quality. \textit{OWA-QP} performs slightly better, but cannot scale past $5$ scenarios, highligting the importance of the proposed Moreau envelope smoothing technique (Section \ref{subsec:moreau_solver}).  

For comparison, grey bars represent the result due to a non-smoothed OWA LP \eqref{model:OWA_LP_form} implemented with implicit differentiation in \texttt{cvxpylayers} \citep{agrawal2019differentiable}. Its poor performance under the OWA subgradient training shows that the proposed approximations of Section \ref{sec:diff_owa_opt} are indeed actually necessary for accurate training.

Runtimes of the smoothed models \eqref{model:OWA_LP_smooth} and \eqref{model:OWA_moreau_solver} are compared in Figure \ref{fig:portfolio_runtimes} (Appendix \ref{app:portfolio_runtimes}). These results show that the Moreau envelope smoothing maintains low runtimes as $m$ increases, while the QP approximation suffers past $m=5$ and causes memory overflow past $m=6$.





\vspace{-6pt}
\subsubsection{Surrogate Learning for OWA Optimization: Multi-species Warcraft Shortest Path}
\label{sec:experiments_warcraft}
This experiment serves to illustrate how a surrogate model can assist in end-to-end training \eqref{eq:OWA_PtO_ERM} when the full OWA problem \eqref{eq:param_OWA_general} is too difficult to solve. Warcraft Shortest Path (WSP) is a popular dataset for benchmarking PtO methods \citep{vlastelica2020differentiation,berthet2020learning}, in which observable features $\bm{z}$ are RGB images of $12 \times 12$ tiled Warcraft maps. A character's movement speed depends on the terrain type of each tile, and the goal is to predict the node-weighted shortest path from top-left to bottom-right where nodes are tiles and weights depend on movement speed. 

This experiment is a multi-objective variation inspired by \citep{botang23}, where multiple species have \emph{distinct node weights} based on their movement speeds   on each terrain type, and must traverse each map together by a \emph{single path}. So that all species travel together as fast as possible, we aim to minimize their OWA-aggregated path lengths.

Noting that node weights can readily be converted to edge weights, the shortest path problem as a linear program reads 
\begin{equation}
    \label{model:LP_shortest_path}
    \bm{x}^{\star}(\bm{c}) = {\argmax}_{ \bm{A}\bm{x} = \bm{b}, \; \bm{0} \leq \bm{x} \leq \bm{1}} \;\; -\bm{c}^T \bm{x}  
\end{equation}
where $\bm{A}$ is a graph incidence matrix, $\bm{b}$ indicates source and sink nodes, and $\bm{c}$ holds the graph's edge weights. A classic result states that due to total unimodularity in $A$, solutions $\bm{x}^{\star}$ to \eqref{model:LP_shortest_path} are guaranteed to take on integer values, so that $\bm{x}^{\star}(\bm{c}) \in \{ 0,1 \}^n$ form valid paths \citep{cormen2022introduction}.

Replacing the linear objective of \eqref{model:LP_shortest_path} with an OWA aggregation over $\bm{C}\bm{x}$ (where rows of $\bm{C} \in \mathbb{R}^{m \times n}$ hold edge weights per species) breaks this property, so that additional integer constraints $\bm{x} \in \{ 0,1 \}^n$ are required, leading to an intractable OWA integer program:
\begin{equation}
    \label{model:OWA_shortest_path}
    \bm{x}^{\star}(\bm{C}) = {\argmax}_{ \bm{A}\bm{x} = \bm{b}, \;  \bm{x} \in \{ 0,1 \}^n} \;\; \textsc{OWA}_{\bm{w}}(-\bm{C} \bm{x})
\end{equation}
Rather than training a predictor of $\hat{\bm{C}} \in \mathbb{R}^{m \times n}$ together with \eqref{model:OWA_shortest_path}, we predict $\hat{\bm{c}} \in \mathbb{R}^{n}$ with \eqref{model:LP_shortest_path} as a differentiable LP surrogate model using \citep{vlastelica2020differentiation}, along with the OWA aggregated path length $\textsc{OWA}_{\bm{w}}(-\bm{C} \bm{x}^{\star}(\hat{\bm{c}}))$ as a loss function. This ensures integrality of $\bm{x}^{\star}(\bm{c})$ while maintaining an  efficient training procedure which requires only to solve \eqref{model:LP_shortest_path} at each training iteration and at inference.

\begin{figure}
\centerline{\includegraphics[width=0.7\columnwidth]{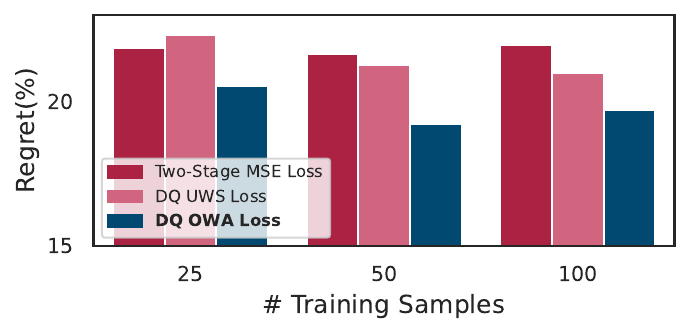}}
\caption{Multi-species OWA path length regret on WSP.}
\label{fig:results_warcraft}
\end{figure}

\paragraph{Settings.}

Three species' node weights are derived by reassigning the movement speeds of each terrain type in the WSP dataset: Humans are fastest on land, Naga on water, and Dwarves on rock, to generate ground-truth $\bm{C} \in \mathbf{R}^{m \times n}$. $\cM_{\theta}$ is a ResNet18 CNN trained to map $12 \times 12$ tiled Warcraft maps to node weights of a shortest path problem \eqref{model:LP_shortest_path}. Blackbox differentiation \citep{vlastelica2020differentiation} is used to backpropagate its solution by Dijkstra's method. See Appendix \ref{app:figures} for an example of the input feature data $\bm{z}$. ResNet18 is strong enough to enable competitive performance by the two-stage given enough data. Following \citep{botang23}, we focus on the limited-data regime, where the advantage of end-to-end learning is best revealed, using $25$, $50$ or $100$ training samples and $1000$ for testing. 

\paragraph{Results.}
Figure \ref{fig:results_warcraft} records percentage regret due to two-stage and unweighted sum baseline models, along with the proposed differentiable LP surrogate trained under OWA loss. Notice how, in each case, the OWA-trained model shows a clear advantage in minimization of OWA regret. 

Table \ref{table:regret_per_species} (Appendix \ref{app:figures}) shows the per-species regret in terms of path length, and reveals that OWA training significantly imrpoves the highest path-length among species, which is intuitive to provide fairness and the main contributor to the aggregated OWA value.

\subsection{Nonparametric OWA with Blackbox Solver: Fair Learning to Rank}
\label{sec:fair_ltr}

The final application setting studies the fair learning-to-rank problem, in which a prediction model ranks $n$ web search results in order of relevance to a user query, while maintaining fairness of exposure across protected groups within the search results. The proposed model learns relevance scores $\bm{c}$ end-to-end with a fair ranking optimization module:
\begin{equation}
    \label{model:OWA_fair_rank}
    \bm{\Pi}^{\star}(\bm{c}) = {\argmax}_{\bm{\Pi}  \in \cB} \;\; 
     (1 - \lambda) \cdot \bm{c}^T\bm{\Pi} \, \bm{b} +  \lambda \cdot  \textsc{OWA}_{\bm{w}}(\cE_G (\bm{\Pi})),
\end{equation}
wherein $\cB$ is the set of all bistochastic matrices, $\bm{\Pi} \in \mathbb{R}^{n \times n}$ represents a ranking policy whose $(i,j)^{\textit{th}}$ element is the probability  item $i$ takes position $j$ in the ranking, $\bm{c}$ measures relevance of each item to a user query, $\bm{b}$ are position bias factors measuring the exposure of each ranking position, and $\bm{c}^T\bm{\Pi} \, \bm{b}$ is the expected Discounted Cumulative Gain, a common measure of user utility. This primary objective is combined with OWA aggregation of the exposure vector $\cE_G (\bm{\Pi})$, whose elements $\cE_g (\bm{\Pi}) =  \bm{1}_g^T \bm{\Pi} \, \bm{b}$ measure the exposures attained by each of several protected groups $g \in G$ where $\bm{1}_g$ hold binary indicators of item inclusion in group $g$. The factor $\lambda$ controls a tradeoff between user utility and group fairness of exposure. 

Since $\bm{b}$ and $\bm{1}_g$ in $\cE_G (\bm{\Pi})$ are known and not modeled parametrically, the problem \eqref{model:OWA_fair_rank} is an instance of \eqref{model:OWA_opt_nonparametric} and its SPO+ subgradient can be modeled as per Section \ref{subsec:nonparametric_OWA_LP}. Solutions $\bm{\Pi}^{\star}(\bm{c})$ are obtained for any $\bm{c}$ by an adaptation of the Frank-Wolfe method with smoothing proposed in \citep{do2022optimizing}, as detailed in Appendix \ref{app:fair_learning_to_rank}.

\paragraph{Settings.}
A feedforward network $\cM_{\theta}$ is trained to predict for $n$ items, given features $\bm{z}$, their relevance scores $\bm{c} \in \mathbb{R}^n$. The SPO+ training scheme of Section \ref{subsec:nonparametric_OWA_LP} is used to minimize regret in \eqref{model:OWA_fair_rank} due to error in $\hat{\bm{c}} = \cM_{\theta}(\bm{z})$. The Microsoft Learning to Rank (MSLR) dataset is used, where $\bm{z}$ are features of items to be ranked and $\bm{c}$ are their relevance scores. Protected item groups are assigned as evenly spaced quantiles of its Quality Score feature.   Each method is evaluated on the basis of mean utility $\bm{c}^T \bm{\Pi} \, \bm{b}$ and fairness violation $\frac{1}{|G|} \sum_{g \in G} \left| \frac{1}{n} \bm{1}^T \cE_G (\bm{\Pi}) -  \cE_g (\bm{\Pi}) \right|$, and their relative tradeoffs over the full range of its fairness parameter.

\paragraph{Baseline Models.}
The model proposed in this section is called Smart OWA Optimization for Fair Ranking (SOFaiR). Selected baseline methods from the fair learning to rank domain include FULTR \citep{singh2019policy}, DELTR \citep{zehlike2020reducing}, and SPOFR \citep{kotary2022end}, futher details are provided in Appendix \ref{app:fair_learning_to_rank}.

\begin{figure}[!ht]
    \centering
    \includegraphics[width=0.45\textwidth]{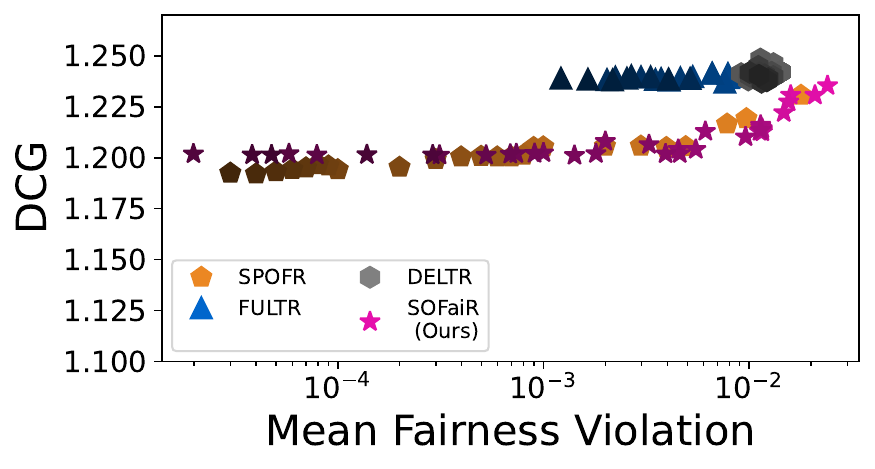} \hfill
    \includegraphics[width=0.45\textwidth]{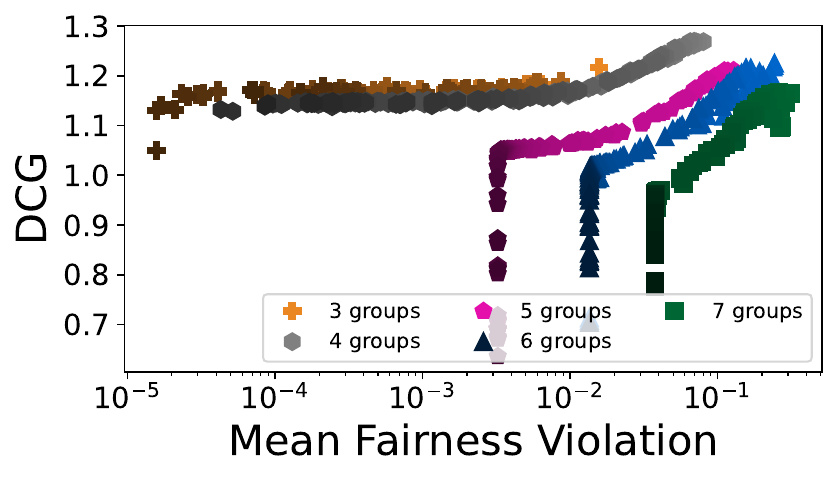}
    \caption{Fairness-utility tradeoffs on MSLR. Left: binary group case, right: multi-group case}

    \label{fig:binary_trade_off}
\end{figure}

\paragraph{Results.}
Figure \ref{fig:binary_trade_off} (left) shows that by enforcing fairness via an embedded optimization, SoFaiR achieves order-of-magnitude lower fairness violations than FULTR or DELTR, which rely on loss function penalties to drive down violations. However, it is Pareto-dominated over a small regime by those methods. Its fairness-utility tradeoff is comparable to SPOFR, which also uses constrained optimization. Notably though, SoFaiR demonstrates order-of-magnitude runtime advantages over SPOFR in Appendix \ref{app:fair_learning_to_rank}. 

Figure \ref{fig:binary_trade_off} (right) shows the analogous result over datasets with $3$-$7$ protected groups. None of the baseline methods are equipped to handle multiple groups on this dataset, but SOFaiR accomodates more groups naturally by OWA optimization over their expected group exposures.

\section{Related Work}
\label{sec:related_work}
Modern approaches to the Predict-Then-Optimize setting, formalized in Section \ref{subsec:predict-then-optimize}, typically maximize decision quality as a loss function, enabled by backpropagation through the mapping $\bm{c} \to \bm{x}^{\star}(\bm{c})$ defined by \eqref{eq:opt_generic}. When this mapping is differentiable, backpropagation can be performed using differentiable optimization libraries \citet{amos2017optnet,agrawal2019differentiable,agrawal2019differentiating,kotary2023folded}.

However, many important classes of optimization are nondifferentiable, including linear and mixed-integer programs.  Effective training techniques  are typically based on forming continuous approximations of \eqref{eq:opt_generic}, whether by smoothing the objective function \citet{amos2019limited,wilder2018melding,mandi2020interior}, introducing random noise \citet{berthet2020learning, paulus2020gradient}, or estimation by finite differencing \citet{poganvcic2019differentiation}.  This paper falls into that category, due to nondifferentiability of the OWA objective, requiring approximation of \eqref{eq:opt_generic} by differentiable functions. 

\section{Conclusions}
This work has presented a comprehensive methodology for incorporating Fair OWA optimization end-to-end with predictive modeling. Beginning with novel differentiable approximations to OWA programs, its proposed toolset also included special techniques to exploit problem specific structure arising in practical problems, such as nonparametric OWA objectives and totally unimodular constraints. These developments were used to demonstrate the potential of Fair OWA optimization in data-driven decision making, with results not previously possible on important problems, such robust resource allocation and fair learning to rank. 

We believe that this work could pave the way to the use of Fair OWA in learning pipelines to enable an array of important multi-optimization problems in many enegineering domains.

\bibliography{arxiv_main}

\appendix

\section{Portfolio Optimization Experiment}
\label{app:portfolio}

\subsection{Efficiency of Differentiable OWA Solvers}

\label{app:portfolio_runtimes}

\begin{figure}
\centering
\includegraphics[width=0.5\columnwidth]{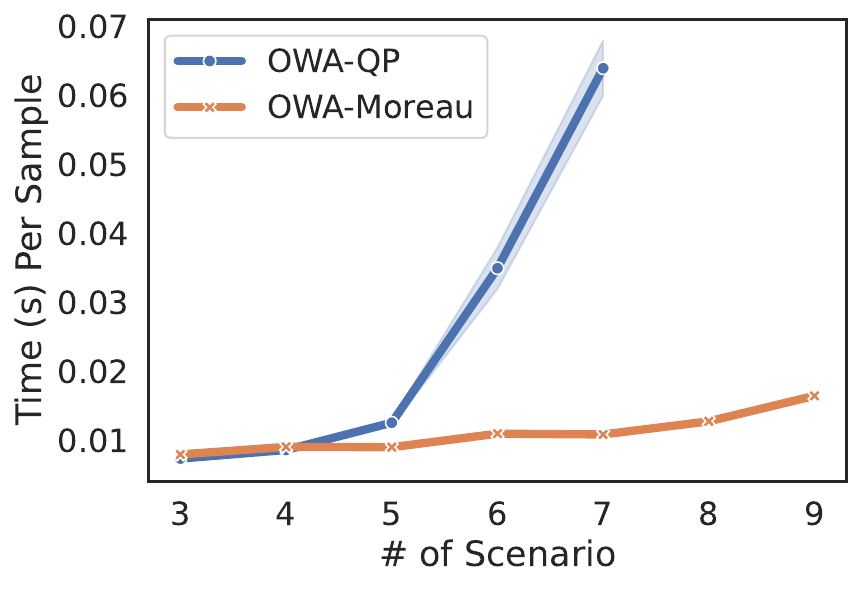}
\caption{Average solving time of 2 smoothed OWA optimization models, on Robust Portfolio Optimization, over $1000$ input samples. Missing datapoints past $7$ scenarios are due to memory overflow as the QP model grows factorially.}
\label{fig:portfolio_runtimes}
\end{figure}

Figure \ref{fig:portfolio_runtimes} depicts the running times of two differentiable optimization models applied to the Portfolio problem. It is evident that for the OWA-LP model, the running time scales factorially with the number of scenarios due to the number of constraints, while for the OWA-Moreau model, it scales linearly. It is noteworthy that the OWA-LP model cannot run with more than 7 scenarios due to memory constraints (requiring over 300GB+).

\subsection{Effect of adding MSE loss}
\begin{figure}
    \centering
\includegraphics[width=0.32\textwidth]{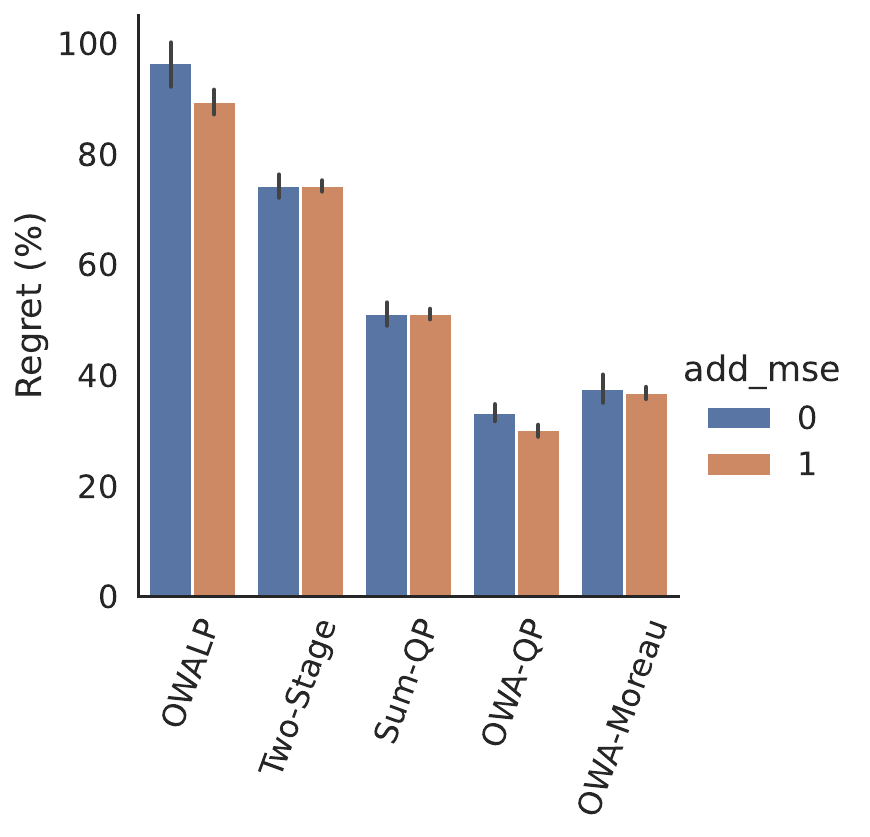}    
  \includegraphics[width=0.32\textwidth]{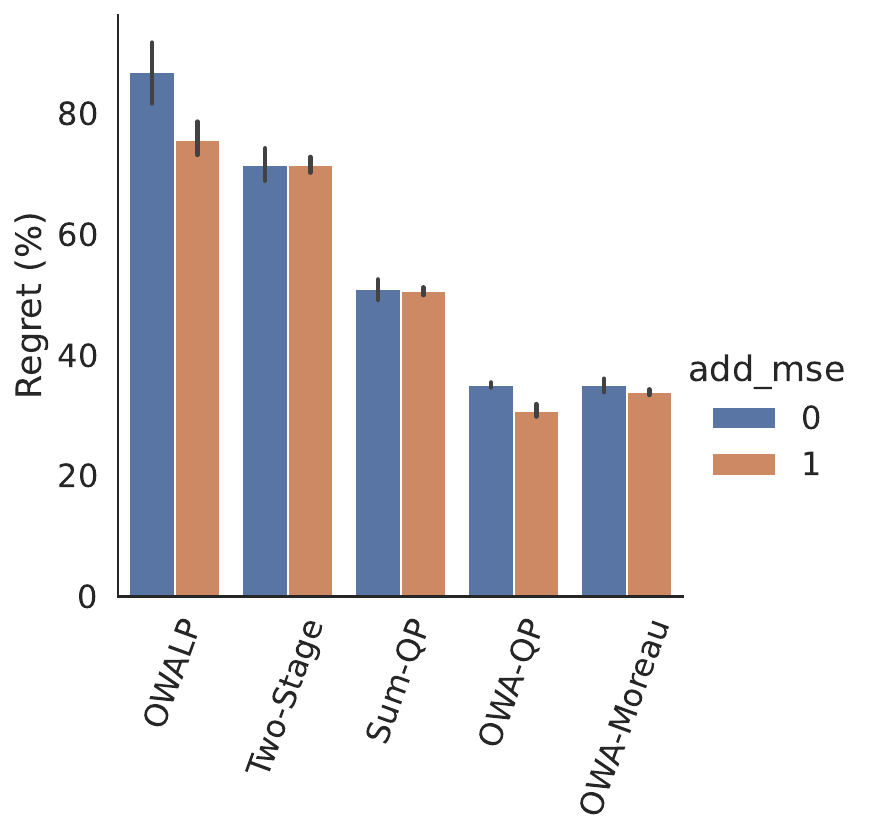}      
\includegraphics[width=0.32\textwidth]{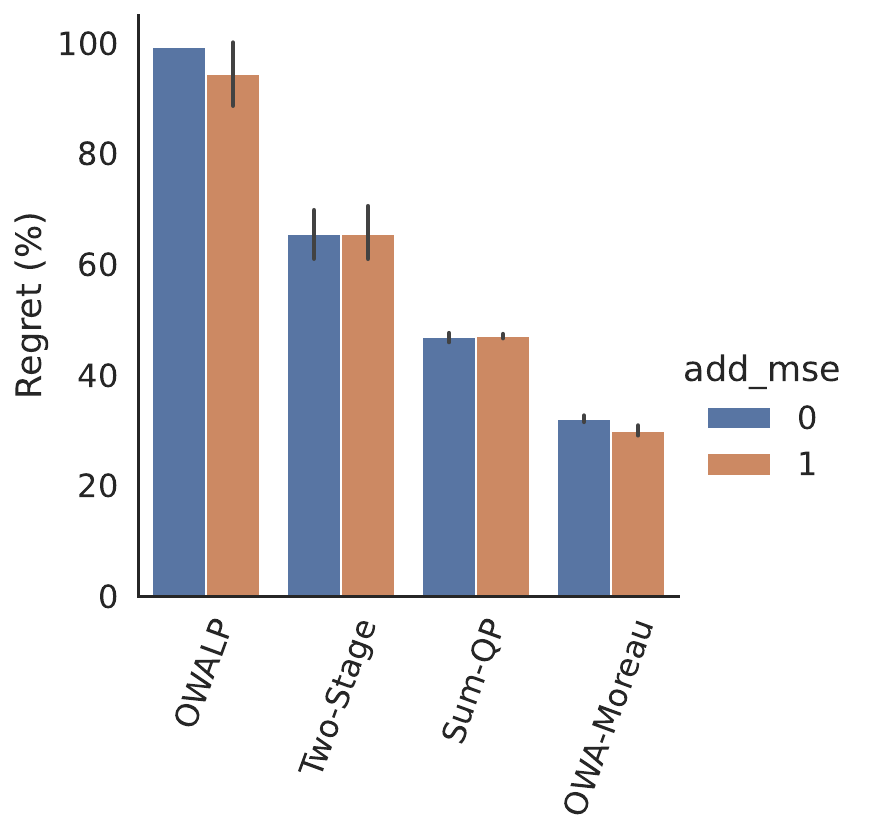}    

    \caption{Effect of MSE Loss on differentiable optimization models. From left to right: 3, 5, 7 scenarios }
    \label{fig:mse}
\end{figure}

Figure \ref{fig:mse} illustrates the impact of combining the Mean Squared Error loss $\cL_{MSE}$ in a weighted combination with the decision quality loss $\cL_{DQ}$. With the exception of OWA-LP, which exhibited instability, and Two-Stage, already trained with MSE Loss, the addition of MSE resulted in slight enhancements to the regret performance.

\subsection{Models and Hyperparameters}

A neural network (NN) with three shared hidden layers following by one separated hidden layer for each species is trained using Adam Optimizer and with a batch size of 64. The size of each shared layer is halved, the output dimension of the separated layer equal to the number of assets. Hyperparameters were selected as the best-performing on average among those listed in Table \ref{tab:hyperparams_portfolio}).  
Results for each hyperparameter setting are averaged over five random seeds. In the OWA-Moreau model, the forward pass is executed using projected gradient descent for 300, 500, and 750 iterations, respectively, for scenarios with 3, 5, and 7 inputs. The update step size is set to $\gamma = 0.02$.

\begin{table}[tb]
\centering
  \caption{Hyperparameters}
\begin{tabular}{rl lllllll}
\toprule
  Hyperparameter    & \multicolumn{1}{c}{Min} 
  & \multicolumn{1}{c}{Max} & 
  \multicolumn{4}{c}{Final Value} \\
  \cmidrule(r){4-9} 
  & & &  OWA-LP & Two-Stage & Sum-QP & OWA-QP & OWA-Moreau & Sur-QP\\
  \midrule
  learning rate   & $1e^{-3}$ & $1e^{-1}$ &\bm{$1e^{-2}$}&$\bm{5e^{-3}}$ & $\bm{1e^{-2}}$ & $\bm{1e^{-2}}$ & $\bm{1e^{-2}}$& $\bm{1e^{-2}}$ \\ 
  smoothing parameter $\epsilon$ & 0.1 & 1.0 & \textbf{N/A} & \textbf{N/A} & 1.0 & 1.0 &\textbf{N/A} & 1.0 \\
  smoothing parameter $\beta_0$ & 0.005 & 10.0 & \textbf{N/A}  & \textbf{N/A} & \textbf{N/A} & \textbf{N/A} &0.05 &\textbf{N/A} \\
   MSE loss weight $\lambda$  & 0.1 & 0.5 & 0.4  & \textbf{N/A} & 0.3 & 0.4 & 0.1 & 0.3 \\
  \bottomrule
\end{tabular}
\label{tab:hyperparams_portfolio}
\end{table} 

\subsection{Solution Methods}
\label{app:portfolio_solver}
The OWA portfolio optimization problem \eqref{eq:param_OWA_portfolio} is solved at test time, for each compared method, by projected subgradient descent using OWA subgradients \eqref{eq:owa_subgrad} and an efficient projection onto the unit simplex $\Delta$ as in \cite{martins2016softmax}:

\begin{equation}
    \bm{x}^{k+1} = \proj_{\Delta} \left( \bm{x}^{k} - \alpha \frac{\partial}{\partial \bm{x}} \textsc{OWA}_{\bm{w}}(\bm{C} \bm{x}) \right)
\end{equation}

For the Moreau-envelope smoothed OWA optimization \eqref{model:OWA_moreau_solver} proposed for end-to-end training, the main difference is that its objective function is differentiable (with gradients \eqref{eq:OWA_moreau_grad}), which allows solution by a more efficient Frank-Wolfe method \cite{beck2017first}, whose inner optimization over $\Delta$ reduces to the simple argmax function which returns a binary vector with unit value in the highest vector position and 0 elsewhere, which can be computed in linear time:

\begin{equation}
    \bm{x}^{k+1} = \frac{k}{k+2 } \bm{x}^{k}  + \frac{2}{k+2 } \argmax \left( \frac{\partial}{\partial \bm{x}} \textsc{OWA}_{\bm{w}}(\bm{C} \bm{x}^k)  \right)
\end{equation}

\section{Fair Learning to Rank Experiment}
\label{app:fair_learning_to_rank}

\subsection{Fair Ranking Optimization by Frank-Wolfe with Smoothing}

This section explains the adaptation of a Frank-Wolfe method with objective smoothing, due to \cite{do2022optimizing}, to solve the fair ranking optimization mapping \eqref{model:OWA_fair_rank} proposed for end-to-end fair learning to rank in this paper.

Frank-Wolfe methods solve a convex constrained optimization problem $\argmax_{\mathbf{x} \in \mathbb{S}} f(\mathbf{x}) $ by computing the iterations 
\begin{equation}
    \label{eq:frank_wolfe_generic}
        \mathbf{x}^{(k+1)} = (1 - \alpha^{(k)})\mathbf{x}^{(k)} + \alpha^{(k)} \argmax_{\mathbf{y} \in \mathbb{S}} \langle \mathbf{y}, \nabla  f(\mathbf{x}^{(k)}) \rangle.
\end{equation}
Convergence to an optimal solution is guaranteed when $f$ is \emph{differentiable} and with $\alpha^{(k)} = \frac{2}{k+2}$ \citep{beck2017first}. However, the main obstruction to solving \eqref{model:OWA_fair_rank} by the method \eqref{eq:frank_wolfe_generic} is that $f$ in our case includes a \emph{non-differentiable} OWA function. A path forward is shown in \citep{lan2013complexity}, which shows convergence can be guaranteed by optimizing a smooth surrogate function $f^{(k)}$ in place of the nondifferentiable $f$ at each step of \eqref{eq:frank_wolfe_generic}, in such a way that the $f^{(k)}$ converge to the true $f$ as $k \to \infty$.

It is proposed in \citep{do2022optimizing} to solve a two-sided fair ranking optimization with OWA objective terms, by the method of \citep{lan2013complexity}, where $f^{(k)}$ is chosen to be a Moreau envelope $h^{\beta_k}$ of $f$, a $\frac{1}{\beta_k}$-smooth approximation of $f$ defined as \citep{beck2017first}:
\begin{equation}
    h^{\beta}(\mathbf{x}) = \min_{\mathbf{y}} f(\mathbf{y}) + \frac{1}{2 \beta} \|   \mathbf{y} - \mathbf{x}  \|^2.
\end{equation}
When $f = \textsc{OWA}_{\bm{w}}$, let its Moreau envelope be denoted $\nabla \textsc{OWA}_{\bm{w}}^{\beta}$;
it is shown in \citep{do2022optimizing} that its gradient can be computed as a projection onto the permutahedron induced by modified OWA weights $\tilde{\bm{w}} = -(w_m, \ldots, w_1)$. By definition, the permutahedron $\cC(\tilde{\bm{w}}) = \textsc{conv}(\{ \bm{w}_{\sigma}: \forall \sigma \in \cP_m \})$ induced by a vector $\tilde{\bm{w}}$ is the convex hull of all its permutations. In turn, it is shown in \citep{blondel2020fast} that the permutahedral projection $\nabla \textsc{OWA}_{\bm{w}}^{\beta}(\mathbf{x}) = \proj_{\cC(\tilde{\bm{w}})} (\nicefrac{\bm{x}}{\beta})$ can be computed in $m \log m$ time as the solution to an isotonic regression problem using the Pool Adjacent Violators algorithm. To find the overall gradient of $\textsc{OWA}_{\bm{w}}^{\beta}$ with respect to optimization variables $\mathbf{\Pi}$, a convenient form can be derived from the chain rule:
\begin{equation}
    \nabla_{\bm{\Pi}} \; \textsc{OWA}_{\bm{w}}^{\beta}(\cE(\bm{\Pi}))= \bm{\mu} \bm{b}^T.
\end{equation}
where $\bm{\mu} = \proj_{\cC(\tilde{\bm{w}})} (\nicefrac{\cE(\mathbf{\Pi})}{\beta})$ and $\cE(\bm{\Pi})$ is the vector of all item exposures \citep{do2022optimizing}. For the case where group exposures $\cE_G(\bm{\Pi})  \cE_g(\mathbf{\Pi}) = \bm{1}_g^T \mathbf{\Pi} \, \bm{b}$ are aggregated by OWA, $\cE_G(\bm{\Pi}) = \bm{A} \bm{\Pi} \bm{b}$, where $\bm{A}$   is the matrix composed of stacking together all group indicator vectors $\bm{1}_g \; \forall g \in G$. Since $\cE(\bm{\Pi}) = \bm{\Pi} \bm{b}$, this implies   $\cE_G(\bm{\Pi}) = \cE(\bm{A} \bm{\Pi})$, thus 
\begin{equation}
    \nabla_{\bm{\Pi}} \; \textsc{OWA}_{\bm{w}}^{\beta}(\cE_G(\bm{\Pi}))= (\bm{A}^T \tilde{\bm{\mu}}) \; \bm{b}^T.
\end{equation}
by the chain rule, and where $\tilde{\bm{\mu}} = \proj_{\cC(\tilde{\bm{w}})} (\nicefrac{\cE_G(\bm{A} \mathbf{\Pi})}{\beta})$. It remains now to compute the gradient of the user relevance term $u(\bm{\Pi}, \hat{\bm{y}}_q) =  \hat{\bm{y}}_q^T \bm{\Pi} \; \bm{b}$ in Problem \ref{model:OWA_fair_rank}. As a linear function of the matrix variable $\bm{\Pi}$, its gradient is $\nabla_{\bm{\Pi}} \; u(\bm{\Pi}, \hat{\bm{y}}_q) =  \hat{\bm{y}}_q \; \bm{b}^T$, which is evident by comparing to the equivalent vectorized form $ \hat{\bm{y}}_q^T \bm{\Pi} \; \bm{b} = \overrightarrow{ \hat{\bm{y}}_q \; \bm{b}^T} \cdot \overrightarrow{\bm{\Pi}}$. Combining this with \eqref{eq:OWA_moreau_grad}, the total gradient of the objective function in \eqref{model:OWA_fair_rank} with smoothed OWA term is  $(1 - \lambda) \cdot \hat{\bm{y}}_q \; \bm{b}^T  +  \lambda \cdot     (\bm{A}^T\tilde{\bm{\mu}}) \; \bm{b}^T $, which is equal to $\left((1 - \lambda) \cdot  \hat{\bm{y}}_q +  \lambda \cdot (\bm{A}^T\tilde{\bm{\mu}}) \right)\; \bm{b}^T$. Therefore the SOFaiR module's Frank-Wolfe linearized subproblem is
\begin{equation}
    \label{eq:sofair_subproblem}
    \argmax_{\bm{\Pi} \in \cB}  \left\langle \bm{\Pi}, \left((1 - \lambda) \cdot  \hat{\bm{y}}_q +  \lambda \cdot (\bm{A}^T\tilde{\bm{\mu}}) \right)\; \bm{b}^T   \right\rangle
\end{equation}

\begin{algorithm}[!t]
\caption{Frank-Wolfe with Moreau Envelope Smoothing to solve \eqref{model:OWA_fair_rank}}\label{alg:FWS}
\KwInput{predicted relevance scores $\bm{\hat{y}} \in \RR^n$, group mask $\bm{A}$, max iteration T, smooth seq. $(\beta_k)$}
\KwOutput{ranking policy $\bm{\Pi}^{(T)} \in \RR^{ n \times n}$}

Initialize $\bm{\Pi}^{(0)}$ as $\bm{P} \in \cP$ which sorts $\bm{{\hat{y}}}$ in decreasing order\;
\For{$k = 1, \ldots, T$}
{
    $\tilde{\bm{\mu}} \gets \proj_{\cC(\tilde{\bm{w}})} (\nicefrac{\cE_G(\bm{A} \mathbf{\Pi})}{\beta_k})$\;
    
    $\hat{\bm{\mu}} \gets (1 - \lambda) \cdot  \hat{\bm{y}}_q +  \lambda \cdot (\bm{A}^T\tilde{\bm{\mu}}) $\;
    
    $\hat{\sigma} \gets \textit{argsort} (-\hat{\bm{\mu}})$\;
    
    Let $\bm{P}^{(k)} \in \mathcal{P}$ such that $\bm{P}^{(k)}$ represents $\hat{\sigma}$\;
    
    $\bm{\Pi}^{(k)} \gets \frac{k}{k+2} \bm{\Pi}^{(k-1)} + \frac{2}{k+2} \bm{P}^{(k)} $\;
}
Return $\bm{\Pi}^{(T)}$\;
\end{algorithm}

To implement the Frank-Wolfe iteration \eqref{eq:frank_wolfe_generic}, this linearized subproblem should have an efficient solution. To this end, the form of each gradient above as a cross-product of some vector with the position biases $\bm{b}$ can be exploited. Note that as the expected DCG under relevance scores $\bm{y}$, the function $\bm{y}^T \Pi \; \bm{b}$ is maximized by the permutation matrix $\bm{P} \in \cP_n$ which sorts the relevance scores $\bm{y}$ decreasingly. But since $ \bm{y}^T \bm{\Pi} \; \bm{b} = \overrightarrow{ \bm{y} \; \bm{b}^T} \cdot \overrightarrow{\bm{\Pi}}$, we identify $ \bm{y}^T \bm{\Pi} \; \bm{b}$ as the linear function of $\overrightarrow{\bm{\Pi}}$ with gradient $\overrightarrow{ \bm{y} \; \bm{b}^T}$. Therefore problem \eqref{eq:sofair_subproblem} can be solved in $\cO(n \log n)$, simply by finding $\bm{P} \in \cP_n$ as the argsort of the vector $ ((1 - \lambda) \cdot  \hat{\bm{y}}_q +  \lambda \cdot (\bm{A}^T\tilde{\bm{\mu}}) )$ in decreasing order. A more formal proof, cited in \citep{NEURIPS2021_48259990}, makes use of \citep{hardy1952inequalities}.

\subsection{Running Time Analysis} 
\label{sec:ranking_running_time}

\begin{figure}
    \centering
    \includegraphics[width=0.7\textwidth]{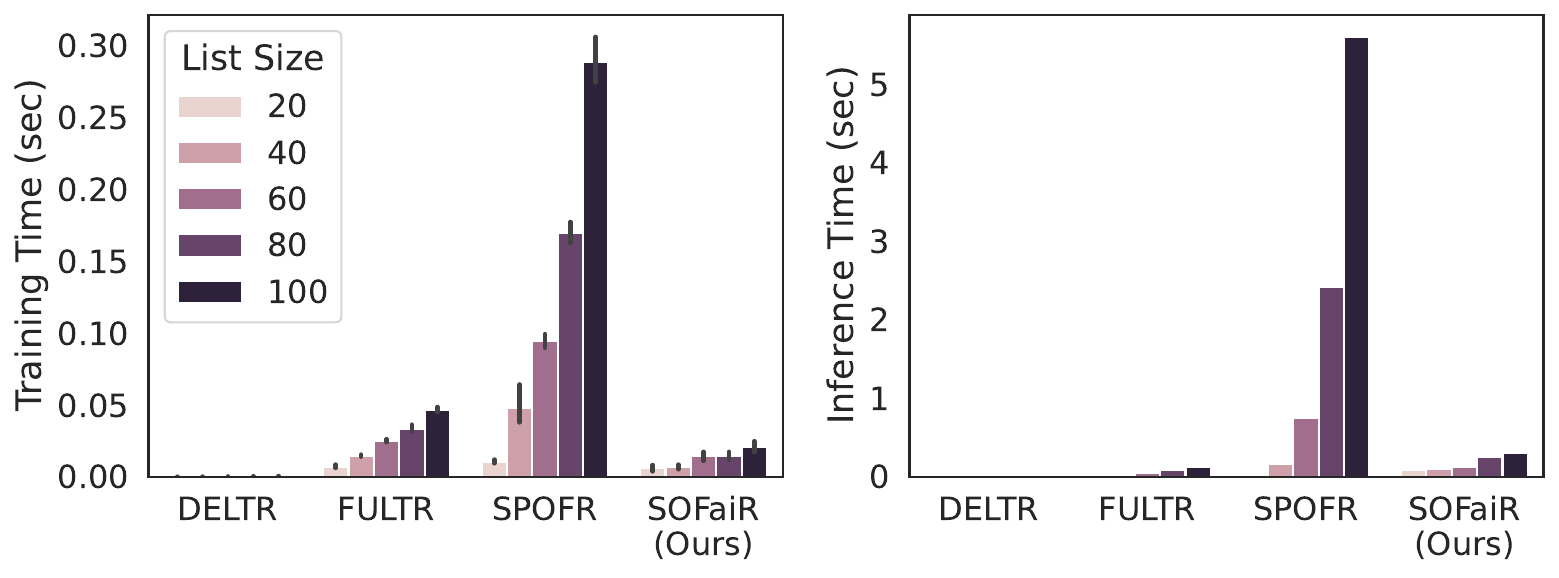}
    \vspace{-8pt}
    \caption{Running time benchmark on MSLR-Web10k dataset}
    \label{fig:benchmark}
\end{figure}

Our analysis begins with a runtime comparison between SOFaiR and other LTR frameworks, to show how it overcomes inefficiency at training and inference time. Figure \ref{fig:benchmark} shows the average training and inference time per query for each method, focusing on the binary group MSLR dataset across various list sizes. First notice the drastic runtime reduction of SOFaiR compared to SPOFR, both during training and inference. While SPOFR's training time exponentially increases with the ranking list size, SOFaiR's runtime increases only moderately, reaching over one order of magnitude speedup over SPOFR for large list sizes. Notably, the number of iterations of Algorithm \ref{alg:FWS} required for sufficient accuracy in training to compute SPO+ subgradients are found to less than those required for solution of \eqref{model:OWA_fair_rank} at inference. Thus the reported results use $100$ iterations in training and $500$ at inference.  Importantly, reported runtimes under-estimate the efficiency gained by SOFaiR, since its PyTorch \citep{paszke2017automatic} implementation in Python is compared against the highly optimized code implementation of Google OR-Tools solver \citep{perron2011operations}. DELTR and FULTR, being penalty-based methods, demonstrate competitive runtime performance. However, this efficiency comes at the expense of their ability to ensure fairness in every generated policy.

\subsection{Models and Hyperparameters}
\smallskip\noindent{\bf Models and hyperparameters.}
A neural network (NN) with three hidden layers is trained using Adam Optimizer with a learning rate of 0.1 and a batch size of 256. The size of each layer is halved, and the output is a scalar item score. Results of each hyperparameter setting is are taken on average over five random seeds.

Fairness parameters, considered as hyperparameters, are treated differently. LTR systems aim to offer a trade-off between utility and group fairness, since the cost of increased fairness results in decreased utility. In DELTR, FULTR, and SOFaiR, this trade-off is indirectly controlled through the fairness weight, denoted as $\lambda$ in \eqref{model:OWA_fair_rank}. Larger values of $\lambda$ indicate more preference towards fairness. In SPOFR, the allowed violation  of group fairness is specified directly. Ranking utility and fairness violation are assessed using average DCG  and fairness violation, respectively. The metrics are computed as averages over the entire test dataset.

\begin{table}[tb]
\centering
  \caption{Hyperparameters}
\begin{tabular}{rl lllll}
\toprule
  Hyperparameter    & \multicolumn{1}{c}{Min} 
  & \multicolumn{1}{c}{Max} & 
  \multicolumn{4}{c}{Final Value} \\
  \cmidrule(r){4-7} 
  & & &  SOFaiR &SPOFR & FULTR & DELTR\\
  \midrule
  learning rate   & $1e^{-5}$ & $1e^{-1}$ &\bm{$1e^{-1}$}&$\bm{1e^{-1}}$ & $\bm{2.5e^{-4}}$ & $\bm{2.5e^{-4}}$ \\ 
  violation penalty $\lambda$   & $1e^{-5}$ & 400 & \textbf{*} & \textbf{N/A} & \textbf{*} & \textbf{*}\\ 
  allowed violation $\delta$    & 0 & 0.01 &\textbf{N/A} & \textbf{*} & \textbf{N/A} & \textbf{N/A}\\ 
  entropy regularization decay  & 0.1 & 0.5 & \textbf{N/A} & \textbf{N/A} & $\bm{0.3}$ & \textbf{N/A}\\ 
  batch size  & 64 & 512 & \textbf{256} & \textbf{256} & \textbf{256}& \textbf{256}\\
  smoothing parameter $\beta_0$ & 0.1 & 100 & \textbf{*} & \textbf{N/A} & \textbf{N/A} & \textbf{N/A} \\
 sample size & 32 & 64&\textbf{N/A} & 64 & 64 & \textbf{N/A} \\
  \bottomrule
\end{tabular}
\label{tab:hyperparams}
\end{table}

Hyperparameters were selected as
the best-performing on average among those listed in Table \ref{tab:hyperparams}. {Final hyperparameters for each model are as stated also in Table \ref{tab:hyperparams}, and Adam optimizer is used in the production of each result.} Asterisks (*) indicate that there is no option for a final value, as all values of each parameter are of interest in the analysis of fairness-utility tradeoff, as reported in the experimental setting Section.

For OWA optimization layers, $\bf{w}$ is set as $w_j = \frac{n-1+j}{n}$, $T=100$ during training , and $T=500$ during testing.

\subsection{Additional Results}
This section includes additional results for fair learning to rank on MSLR, in which list sizes to be ranked are increased to $100$ items. This allows runtimes to be compared as a function of list size, which determines the size of the fair ranking optimization problem. It also reveals how penalty-based methods DELTR and FULTR suffer in their ability to satisfy fairness accurately.

\begin{figure}
    \centering
    \includegraphics[width=0.45\textwidth]{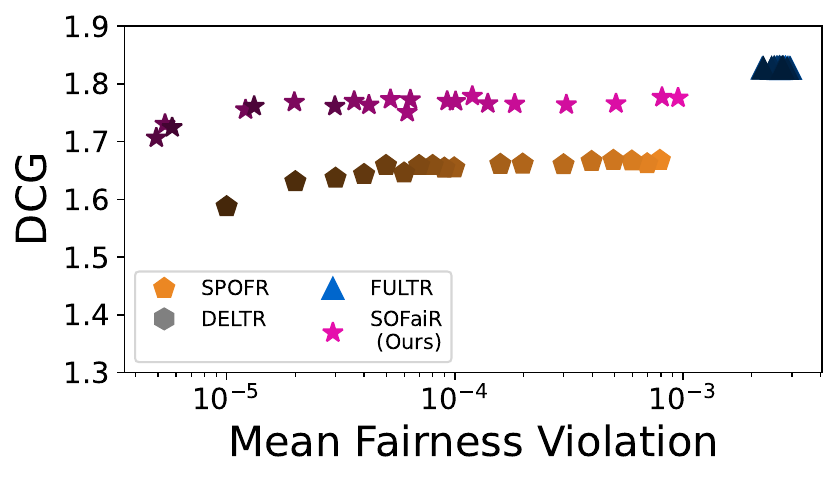}
    \caption{Fairness-utility tradeoffs on MSLR }
    \label{fig:binary_trade_off_100}
\end{figure}

\begin{figure}
    \centering
    \includegraphics[width=0.45\textwidth]{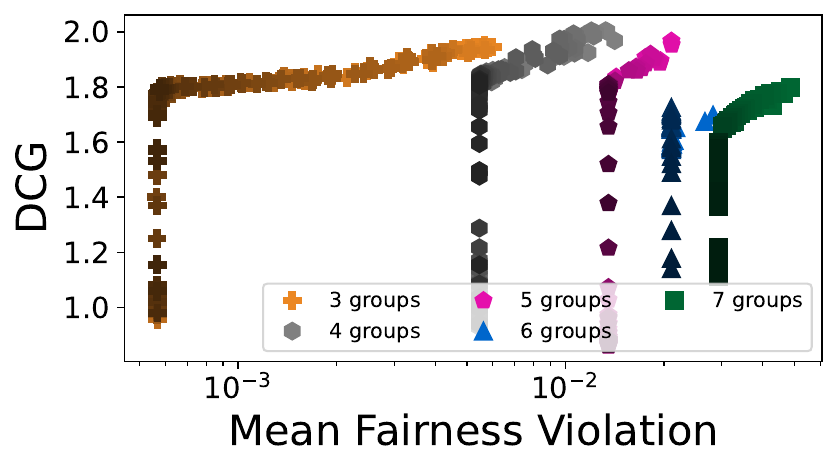}
    \caption{Fairness-utility tradeoffs on MSLR Multi-group}
    \label{fig:multi_trade_off_app}
\end{figure}

\section{Multi-species Warcraft Shortest Path} 
\label{app:figures}
\begin{figure}
    \centering
    \includegraphics[width=0.3\textwidth]{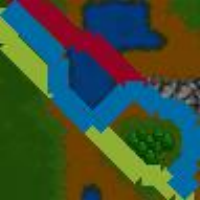}
    \caption{Shortest paths for each species in a Warcraft map}
    \label{fig:shortestpath_map_app}

\end{figure}

Figure \ref{fig:shortestpath_map_app} showcases the Warcraft map featuring the shortest paths for three distinct species. The paths for Humans, Naga, and Dwarves are depicted in green, red, and blue, respectively. Humans excel on land, Naga traverse water most efficiently, while Dwarves navigate rocky terrain with the greatest speed.

Table \ref{table:regret_per_species} presents the regrets for each species across three models with different number of training data. It is notable that the model trained with OWA Loss significantly outperforms the two-stage model by more than 10\% for the Human race. Conversely, the two-stage model exhibits slightly better performance for the Dwarf, albeit by a very small margin (<3\%).

\begin{table*}[!t]
\centering
\resizebox{0.7\linewidth}{!}{
\begin{tabular}{l | rrr | rrr | rrr}
    \toprule
        \multicolumn{1}{c}{\textbf{Model}}     & 
        \multicolumn{3}{c}{Human} &
        \multicolumn{3}{c}{Naga}  &
        \multicolumn{3}{c}{Dwarf}\\
    \cmidrule{2-4}\cmidrule{5-7}\cmidrule{8-10}
    & {25} & {50 } & {100}
    & {25} & {50 } & {100}
    & {25} & {50 } & {100}\\
    \midrule
     Two-Stage MSE Loss  & 44.4 & 44.6 & 46.3 & \bf{34.2} & \bf{34.1} & \bf{33.9} & 44.1 & \bf{41.6}  & \bf{42.8}\\ 
     End2End Sum Loss &51.5  & 49.0 & 47.9 & 35.2 & 33.6 & 34.6 & 43.8 & 43.8 & 43.4\\  
     \textbf{End2End OWA Loss}  & \bf{43.8} & \bf{31.8} & \bf{33.6} & \bf{34.2} & 37.6 & 34.8 & \bf{41.3} & 43.1 & 43.1 \\ 
    \bottomrule
    \end{tabular} 
}
\caption{Regret ($\%$) per species}
\label{table:regret_per_species}
\end{table*}

\end{document}